\documentclass[twoside,11pt]{article}

\usepackage[nohyperref,abbrvbib]{jmlr2e}
\usepackage{amssymb}
\usepackage{amsmath}
\usepackage{booktabs}
\usepackage{tabularx}
\usepackage{colortbl}
\usepackage{xcolor}
\usepackage{array}
\usepackage[colorlinks=true,allbordercolors={1 1 1},citecolor=magenta]{hyperref}
\usepackage{multirow}
\usepackage[inline]{enumitem}

\newcommand{\sd}[1]{\newline\tiny\textcolor{red}{$\pm$ {#1}}}
\newcolumntype{Y}{>{\centering\arraybackslash}X}

\firstpageno{1}

\begin{document}
\sloppy

\title{Towards unstructured mortality prediction\\with free-text clinical notes}

\author{\name Mohammad Hashir \email mohammad.hashir.khan@umontreal.ca \\
       \addr Mila, Universit\'{e} de Montr\'{e}al, Montreal, Canada
       \AND
       \name Rapinder Sawhney \email sawhney@utk.edu \\
       \addr University of Tennessee, Knoxville, USA}

\editor{Kevin Murphy and Bernhard Sch{\"o}lkopf}

\maketitle

\begin{abstract}
Healthcare data continues to flourish yet a relatively small portion, mostly structured, is being utilized effectively for predicting clinical outcomes. The rich subjective information available in unstructured clinical notes can possibly facilitate higher discrimination but tends to be under-utilized in mortality prediction. This work attempts to assess the gain in performance when multiple notes that have been minimally preprocessed are used as an input for prediction. A hierarchical architecture consisting of both convolutional and recurrent layers is used to concurrently model the different notes compiled in an individual hospital stay. This approach is evaluated on predicting in-hospital mortality on the MIMIC-III dataset. On comparison to approaches utilizing structured data, it achieved higher metrics despite requiring less cleaning and preprocessing. This demonstrates the potential of unstructured data in enhancing mortality prediction and signifies the need to incorporate more raw unstructured data into current clinical prediction methods.

\end{abstract}


\section{Introduction}
Mortality is one of the most commonly used outcomes for measuring quality of care in the intensive care unit (ICU) and mortality prediction maintains a substantial presence in practice and research. Predicted mortality can aid in prognostic decision-making by medical professionals and serve as a basis for patient stratification for administrative purposes, billing and research \citep{Russellbook-pfcc4}. 
Severity of disease scoring systems (or severity scores) such as APACHE and SAPS are a widespread method of predicting mortality. These scores are calculated using variables derived from physiological data \& admission attributes and mapped to a probability of mortality using logistic regression. However, the constituent variables can suffer from ambiguous interpretation \citep{Fery-Lemonnier1995-ut}, bias \citep{Suistomaa2000-lv}, loss of temporality through summarization of values and issues in data acquisition due to equipment costs or availability \citep{Haniffa2018-mi}. Severity scores have also demonstrated poor generalizeability across countries \citep{Pappachan1999-ep, Aggarwal2006-ay, Haniffa2018-mi} and diseases \citep{Brown1995-bn, Cheng2017-ma}. The increasing prominence of the electronic health record (EHR) has oriented research in mortality prediction towards more personalized and data-driven machine learning models which seems more worthwhile than adopting standard severity scoring systems \citep{Xie2017-dm}. Machine learning models utilize data which can be the same/similar as severity scores or can be more sophisticated such as clinical time series. 

These methods for predicting mortality usually involve structured data, \textit{i.e.} data that restricts entry to specific fields depending on the type and can usually be represented in a tabular or panel form, and structured data is susceptible to a variety of issues. Similar to severity scores, there can be issues with recording data as equipment can be busy, expensive or prone to error. This can cause data to be missing which is a very common issue and is mitigated with a variety of strategies including imputation. The type of missingness and the strategy used to handle it both may unintentionally introduce bias in the data for downstream tasks \citep{Beaulieu2017-zw}. Another major issue is that healthcare data exists in extensive data warehouses and is often dirty and noisy. Bringing it into a clean structured form requires substantial manual effort in extraction and preprocessing. Moreover, the EHR hosts multiple modalities with widely differing sampling rates which can make joint modeling and analysis significantly laborious. 

Eighty percent of the EHR is composed of unstructured data \citep{MurdochBigH80} which is mostly free-text notes compiled in patient encounters. These notes are a highly untapped resource in clinical support \citep{Shickel2018-py}. Clinician progress notes can contain the most important information about the patient's physiological condition and trajectory \citep{Boag2018-mw}. The unstructured format of the notes allows for recording precise and domain-specific information which can be missed by the structured fields of the EHR \citep{Resnik2008-nt}. As most machine learning models are not designed to work with raw unstructured data, conventional natural language processing (NLP) methods can be applied for extracting structured features. But the onus of extracting clinically relevant information lies on the individual, which could make the feature extraction vulnerable to confirmation bias. 

Deep neural networks recently have demonstrated much success in text classification. These models can work with unstructured data and automate the feature extraction. Moreover, recurrent networks can keep track of temporality. This work examines using raw unstructured notes to predict mortality: a hierarchical approach is employed that exploits the multiple notes collected at different points of time in the patient's stay to model the temporal changes in the patient's state. 

\section{Related work}
Mortality prediction models have traditionally been built with private data and a cohort restricted to a certain ailment(s). With the release of de-identified ICU datasets such as MIMIC-III \citep{johnson2016mimic}, disease-agnostic models and public benchmarks are becoming more common. This section discusses current approaches to mortality prediction specifically on the MIMIC-III dataset, which widely vary in their choice of algorithms, input data and prediction tasks. Plenty of these approaches have been designed to be ``dynamic", i.e. the time of prediction is set at a point during the patient's stay rather than post-discharge. It involves using only the data that had been collected up to the time of prediction, usually the first 24 or 48 hours after admission. This could be due to how severity scores are calculated in the same dynamic manner (24 hours after admission).  

There have been a variety of linear and non-linear models utilized in predicting mortality with the best performing including gradient boosting \citep{johnson2017real}, random forests, SVM \citep{Ghassemi2014-qc}, recurrent neural networks \citep{Harutyunyan2017-hy, Suresh2018-am,Purushotham2017-sf, Bahadori2019-qx}, convolutional neural networks \citep{Grnarova2016-pj, Caicedo-Torres2019-hg}, stacked denoising autoencoders \citep{Sushil2017-hj}, Extreme Learning Machine \citep{Krishnan2018-cq}, hierarchical attention network \citep{Sha2017-ya} and multi-head attention mechanism \citep{Song2018-ag}. The input data to these models has been mostly structured data from the electronic health record (EHR) with the most common being temporal physiological measurements and patient attributes associated with scoring systems such as blood pressure, age, heart rate etc. Other kinds of input data such as prescriptions \& input/output volumes \citep{Purushotham2017-sf}, ICD codes \citep{Sha2017-ya, Ghassemi2014-qc} and notes \citep{Waudby-Smith2018-cm, Kocbek2017-wm, Grnarova2016-pj, Sushil2017-hj, Tran2018-ti, Zalewski2017-vz, Lehman2012-jk, Weissman2018-vj, Krishnan2018-cq, Si2019-ru} have also been used.   While the task of predicting in-hospital mortality has remained prominent across most studies, some works have also predicted post-discharge mortality with periods of 30 days \citep{Purushotham2017-sf, Tran2018-ti, Ghassemi2014-qc, Grnarova2016-pj, Sushil2017-hj} \& one year \citep{Purushotham2017-sf, Ghassemi2014-qc, Grnarova2016-pj, Sushil2017-hj} and also post-admission mortality with periods of two \& three days \citep{Purushotham2017-sf} \& 30 days \citep{Waudby-Smith2018-cm, Kocbek2017-wm}.

Utilizing structured data is widespread but using notes to predict mortality isn't unheard of. Most studies would manually extract structured features from the notes and combine them with other structured data using NLP techniques like topic modeling \citep{Ghassemi2014-qc, Zalewski2017-vz, Lehman2012-jk}, term frequency - inverse document frequency \citep{Kocbek2017-wm}, n-gram statistics \citep{Kocbek2017-wm}, sentiment analysis \citep{Tran2018-ti, Waudby-Smith2018-cm} and bag-of-words \citep{Weissman2018-vj}. 
Some studies have used the raw unstructured notes by applying neural models to map the notes to a meaningful representation for classification.  \citet{Grnarova2016-pj} created a patient representation through two hierarchical convolutional layers at the word and sentence level; this was expanded by \citet{Si2019-ru} to a multi-task learning setup. \citet{Sushil2017-hj} used doc2vec and a stacked denoising autoencoder to calculate the embedding for all the non-discharge notes. \citet{Krishnan2018-cq} used word2vec to summarize ECG reports into a document embedding by summing the vectors of the individual words. 

Studies that are ``dynamic" (only using data from the first \textit{n} hours of the stay) and do account for temporality have used structured data as an input. On the other hand, most of the studies that have exploited the unstructured nature of notes share a single theme: the static and post-hoc nature of the approach. Either only the first note or all the notes (concatenated together) have been used to predict mortality and any time-varying aspects of the patient's state haven't been explicitly modeled.
This work aims to combine the best from both: predicting mortality ``dynamically" with minimally preprocessed unstructured notes while preserving temporality. It tries to learn from the medically rich and relatively less biased contents of notes by using a hierarchical approach, wherein each note is modeled independently yet in a time varying manner.

\section{Model architecture}
Multiple notes are modeled with a hierarchical CNN-RNN (HCR) composed of two modules called the semantical and temporal blocks which consist of convolutional and recurrent layers respectively. The convolutional layers are intended to capture semantic information in individual notes which is then used by the recurrent layers to capture temporal relationships between the notes. The structure of each module is elaborated below:

\begin{itemize}
	\item \textbf{Semantical module}\\It consists of one or more `convolution blocks' which is an abstraction over a sequence of \textit{Conv1D---SpatialDropout---BatchNorm---ReLU}. First, a one-dimensional convolutional layer is applied to an input note which convolves several kernels (also called filters) along the lexical dimension to generate feature maps. This is followed by a special form of dropout called SpatialDropout \citep{Tompson2014-lf} which drops entire feature maps, a batch normalization layer and a ReLU non-linearity. This could be followed by an optional residual connection. The input could be convolved even deeper with additional convolution blocks. Finally, a global average pooling layer is applied to the lexical dimension to generate a vector for the note. 
	\item \textbf{Temporal module}\\It is composed of one or more recurrent layers that keep track of how the condition of the patient has evolved over time using the vectors produced by the previous layer. It produces a vector that represents the state of the patient at the last time step. 
\end{itemize}

The input to the model is a `patient file' consisting of all the preprocessed notes collected from a defined period of the hospital stay and sorted by charting time and a hierarchical approach is used to process it. First, the semantical module is applied to individual notes (the set of weights are shared across all the notes) producing a `document vector' for each notes. The document vectors are then fed sequentially into the temporal block which keeps track of the temporality in the patient's state. The output of the temporal block is a `patient vector' which can then be used for classification. 

Hierarchical models have been used extensively to model text, with hierarchical attention network \citep{Yang2016-cr} being a prominent example. This particular architecture was inspired by these models. Some differences do exist: it forgoes any attention mechanism and replaces the RNN-based encoder at the word-level with a CNN-based encoder to speed up learning. Moreover, there is a document-level encoder in place of a sentence-level encoder to generalize the model to multiple documents without adding another level.

\section{Data and preprocessing}
The MIMIC-III \citep{johnson2016mimic} dataset is used for the analysis. It is a public ICU dataset which comprises information on almost 60,000 critical care admissions at a Boston hospital from 2001-2015. It hosts data from six different types of ICU's and includes vital signs, medications, laboratory measurements, observations notes and more. The adult admissions comprised of 38,597 distinct patients (55.9\% male) with a median age of 65.8 and median length of hospital stay of 6.9 days. There is a significant class imbalance in the dataset as the overall in-hospital mortality is 11.5\%. 

Clinical notes in MIMIC-III are present in the NOTEEVENTS table  and have a variety of authors such as doctors, nurses, imaging professionals, nutritionists and rehabilitation staff. These notes have been de-identified with the identifying information replaced with a unique token following a pattern, for \textit{e.g.} \texttt{[**Hospital1 18**]} denotes a specific de-identified hospital. Two corpuses were created, one for pre-training embeddings and the other for training the prediction model with the latter being a subset of the former. 

First, the NOTEEVENTS table was filtered of any duplicate or erroneous notes and then all the notes were lowercased. Any de-identified names, hospitals and dates in the notes were replaced with a single token \texttt{deidentifiedname}, \texttt{deidentifiedhosp} and \texttt{deidentifieddate} respectively; other types of de-identified tokens were removed. Only tokens containing characters consisting of alphabets, alphabets with numbers (like 24mg) or numbers less than 1000 were kept. Any extra spaces and breaks were reduced to a single space. This formed the first corpus. 

From the first corpus, all the discharge summaries were removed. The remaining notes were truncated/padded up to 500 words to have a uniform data structure as an input. Any notes with missing chart times were given a chart time of midnight of their corresponding chart date. The notes were grouped by hospital stay and sorted by chart time within each group to form the second corpus of `patient files' to be used for training the prediction model. Around 56\% of these notes were written by a physician or a nurse, 39\% were either radiology, echo or ECG reports and the rest were miscellaneous.

\section{Experimental setup}
The proposed approach is compared against two baselines, severity-of-disease scoring systems and structured data. Three severity scores are used in this analysis: Simplified Acute Physiology Score II (SAPS-II), Acute Physiology Score III (APS-III) and Oxford Acute Severity of Illness Score (OASIS). SAPS-II  \citep{le1993new} has been very commonly used in ICU settings, APS-III \citep{johnson2014mortality} is a derivative of APACHE \citep{knaus1991apache} which is another very common scoring system and OASIS \citep{johnson2013new} is a recently developed system from 2014. 

The other baseline used for evaluation are models built with structured data, specifically RNNs using clinical time series termed as CTS-RNN. The feature set used by \citet{Harutyunyan2017-hy} serves as the structured clinical time series for this analysis. It was recreated by replicating their specific extraction and preprocessing paradigm through some modifications to their publicly-hosted code\footnote{github.com/YerevaNN/mimic3-benchmarks}. The cleaned data consisted of 17 types of temporal physiological data such as blood pressure, capillary refill rate, fraction inspired oxygen, Glasgow Coma Scale, glucose, heart rate, height, oxygen saturation, respiratory rate, temperature, weight and pH with additional features indicating missingness.

Two types of HCRs are assessed: a model using only notes called Notes-HCR and a joint model of notes and structured clinical time series called multi-modal HCR or {MM-HCR}. In summary, four types of models are assessed: severity scores, CTS-RNN, Notes-HCR and MM-HCR. All these prediction models were to be `dynamic', \textit{i.e.} they were to be built only with the data from a defined interval in the patient's stay. Experiments were based on the length of the acceptable window of data collection (in hours), referred to as \textit{W}. The starting point of \textit{W} was fixed at the ICU in-time and the endpoint was selected arbitrarily. As severity of disease scoring systems usually use the data only from the first 24 hours, an experiment was defined with $W = 24$. There were two more experiments defined with $W = 12$ and  $W = 48$ to assess how the other three perform with less/more data.

\subsection{Cohort selection}
Cohort selection was done based on \textbf{hospital stays} rather than patients; it is important to make this distinction as a single patient can have multiple hospital stays. The criteria involved filtering out stays that had any of the following attributes
\begin{enumerate*}[label=(\roman*)]
	\item age $\leq$ 18 years
	\item multiple ICU stays in the same hospital stay
	\item transfers between different ICUs and/or wards
	\item time of death within the first 72 hours of their respective ICU stay.
\end{enumerate*}

Experiment-specific cohorts were further created from this initial cohort. The cohort for each experiment consisted of only the hospital stays that had at least one note charted in their respective window. The cohort for the 12 hour, 24 hour and 48 hour windows had a size of 34,207, 35,891 and 36,561 hospital stays respectively. The average number of notes per hospital stay was 3.5, 4.8, and 9.3 for the three windows and in-hospital mortality was approximately 6.5\% in all three.

\subsection{Embedding the notes}\label{fasttext}
The first corpus was used to pre-train word embeddings with fastText \citep{bojanowski2016enriching}. The skip-gram algorithm was run for 100 epochs with an embedding dimension of 200 and a window size of 6. Due to memory constraints, only tokens with more than 20 occurrences in the entire corpus were included in the final vocabulary. All the notes were then embedded into higher dimensional space using these pre-trained embeddings for the input to any HCR.

\subsection{Implementation}
The CTS-RNN baseline was implemented as a two layer deep bi-directional gated recurrent unit (GRU) \citep{Cho2014-py} with hidden sizes 32 \& 16 and weight decay of $10^{-3}$. For the Notes-HCR, the semantical module consisted of three sequential convolution blocks with residual connections and a SpatialDropout probability of 0.5. All convolutions used 200 filters, a kernel size of 3 and weight decay of $10^{-5}$. The temporal module was implemented as a bi-directional GRU with a hidden size of 64. For either CTS-RNN or Notes-HCR, the output vector of the final GRU was fed into a sigmoid layer to get the mortality probabilities. 

\begin{figure}[t]
	\centering     
	\includegraphics[width=0.9\textwidth]{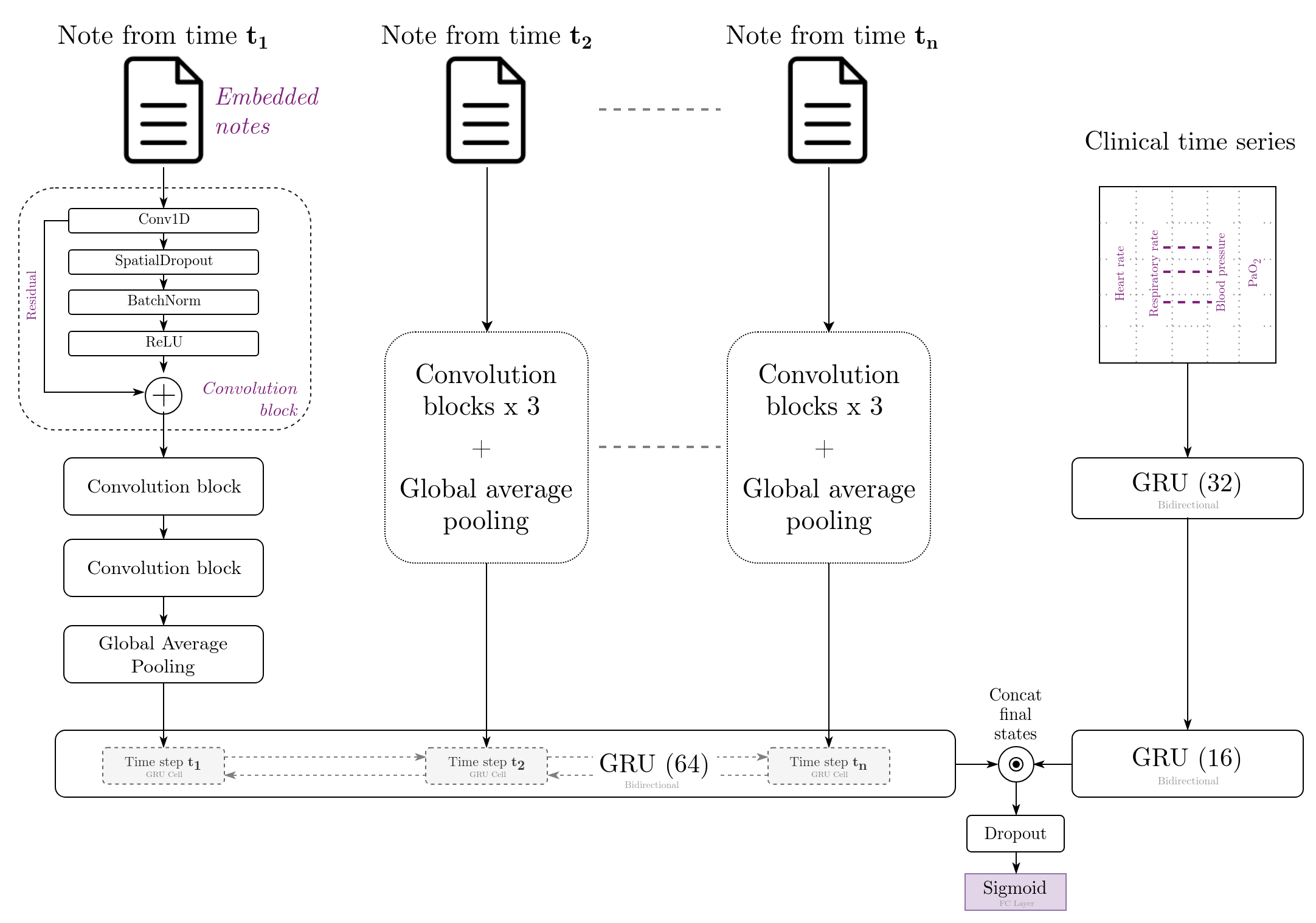}
	\caption{The architecture of the implemented MM-HCR. The semantical module consists of three convolution blocks with residual connections and the temporal module is depicted as an unrolled GRU of state size 64. Two GRUs of state sizes 32 and 16 are used to model the clinical time series.}
	\label{fig:modelarch}
\end{figure}

The MM-HCR used the exact configuration for the CTS-RNN and HCR as described above and just concatenated their respective final GRU's output vectors before the sigmoid. Its architecture is visualized in Figure \ref{fig:modelarch}, also detailing the configuration for the single modality models. A dropout layer with probability of 0.3 was used before the sigmoid in CTS-RNN and MM-HCR. All models were implemented in Keras \citep{chollet2015keras}. 

\subsection{Training}
For a robust evaluation, \textit{k}-fold cross validation with $k=5$ was performed with three folds used as the training set and the remaining two as validation and testing sets. As a single patient could have multiple hospital stays in MIMIC-III, it was ensured that all their hospital stays remained in either the training, validation or testing sets to prevent any information leakage across the sets. 

The binary cross-entropy function was used as the loss function with the AMSGrad \citep{amsgrad2018on} variant of the Adam \citep{Kingma2015-mc} optimizer. As there was a high class imbalance, class weights were used in the loss function. All training runs were for 100 epochs with early stopping. The CTS-RNN model used a batch size of 64 and learning rate of 0.001. Both HCR models used a batch size of 16 and an initial learning rate of 0.001 that was subsequently divided by 10 at epochs 10, 50 and 90. These hyperparameters along with the particular model configurations were found using a manual search.  

The severity scores were calculated using code provided in a public repository by the authors of MIMIC-III\footnote{github.com/MIT-LCP/mimic-code}. This code also calculated the mortality probabilities so fitting of any logistic regression was not required.

\section{Results and discussion}
The performance is evaluated using the area under the ROC curve (AUROC) and the precision-recall curve (AUPRC). The metrics achieved are tabulated in Table \ref{auc}, averaged per fold. All severity scores perform considerably worse than the rest regardless of experiment and metric, with one exception: SAPS-II  performs on par with CTS-RNN on the $W=12$ experiment. The HCR models are better than all severity scores on $W=12$ despite using fewer hours of data. On the $W=24$ experiment, CTS-RNN and both HCRs are immensely better than the severity scores while using the same number of hours of data. The gap between notes and clinical time series reduces significantly on the $W=48$ experiment with only a difference of 0.012 between their average AUROCs. However, a one-tailed t-test found this difference very statistically significant ($p = 0.007$). The AUPRC of notes is significantly higher than clinical time series as well. These results demonstrate the higher discrimination capability of notes on the mortality prediction task.

For all experiments, the joint model MM-HCR achieved the highest average metrics. But the confidence intervals of both of its metrics overlapped with those of Notes-HCR on the $W=12$ and $W=24$ experiments. On performing a one-tailed t-test, the difference between the two models was found to be statistically significant for $W=24$ but not for $W=12$ (for both metrics). The latter result makes sense as the difference in metrics between CTS-RNN and Notes-HCR on $W=12$ is lower than $W=24$. 

\begin{table}[h]
	\centering   \renewcommand{\arraystretch}{1.3}                   
	\caption[Averaged metrics achieved]{Averaged metrics per fold with standard deviation in red. The symbol in superscript for some cells denotes the significance level of the difference from another cell in the  \underline{same column}. $\dagger$ is not statistically significant ($p > 0.05$), * is statistically significant ($p < 0.05$) and ** is very statistically significant ($p < 0.01$)}\label{auc} 
	
	\begin{tabularx}{0.95\textwidth}{lYYYcYYY}
		\arrayrulecolor{black}\toprule[0.2ex]
		\multirow{2}{*}{\textbf{Model}} &            \multicolumn{3}{c}{\textbf{AUROC}}             & &           \multicolumn{3}{c}{\textbf{AUPRC}}             \\
		\arrayrulecolor{black!20}\cmidrule(lr){2-4}						  \cmidrule(lr){6-8}
		& \textit{W}=12     & \textit{W}=24     & \textit{W}=48     & & \textit{W}=12     & \textit{W}=24     & \textit{W}=48     \\  \arrayrulecolor{black}\midrule
		\addlinespace[-1pt]\multicolumn{8}{l}{\scriptsize\textit{Severity scores}} \\[-7pt]
		\text{APS-III}                  & 0.7594\sd{0.0124} & 0.7709\sd{0.0207} & 0.7723\sd{0.0083} & &0.2052\sd{0.0126} & 0.2183\sd{0.0089} & 0.2198\sd{0.0095} \\
		\text{OASIS}                    & 0.7638\sd{0.0077} & 0.7717\sd{0.0107} & 0.7721\sd{0.0050} & &0.2026\sd{0.0121} & 0.2053\sd{0.0075} & 0.2029\sd{0.0071} \\
		\text{SAPS-II}                  & 0.7999\sd{0.0091} & 0.8048\sd{0.0122} & 0.8048\sd{0.0034} & &0.2216\sd{0.0114} & 0.2356\sd{0.0117} & 0.2306\sd{0.0064} \\ \arrayrulecolor{black!20}\midrule 
		\addlinespace[-1pt]\multicolumn{8}{l}{\scriptsize\textit{Single modality}} \\[-7pt]
		\text{CTS-RNN}                  & 0.8090\sd{0.0092} & 0.8425\sd{0.0099} & 0.8765$^{**}$\sd{0.0065} & &0.2342\sd{0.0250} & 0.3010\sd{0.0168} & 0.3475$^{**}$\sd{0.0178} \\
		
		\text{Notes-HCR}                &0.8355$^\dagger$\sd{0.0096} & 0.8675$^{*}$\sd{0.0083} & 0.8887$^{**}$\sd{0.0060} & &0.2753$^\dagger$\sd{0.0197} & 0.3327$^{*}$\sd{0.0116} & 0.3853$^{**}$\sd{0.0156} \\  \arrayrulecolor{black!20}\midrule 
		\addlinespace[-1pt]\multicolumn{8}{l}{\scriptsize\textit{Multi-modal}} \\[-7pt]
		\text{MM-HCR}                   & 0.8463$^\dagger$\sd{0.0107} & 0.8787$^{*}$\sd{0.0061} & \textbf{0.9023}\sd{0.0043} & &0.2985$^\dagger$\sd{0.0256} & 0.3583$^{*}$\sd{0.0207} & \textbf{0.4332}\sd{0.0125} \\ 
		\arrayrulecolor{black}\bottomrule[0.2ex]
	\end{tabularx}

\end{table}

\subsection*{Discussion}\label{sec:discussion}
The severity of diseases scoring systems were developed with expensive studies on cohorts, sometimes spanning multiple countries. The cost of developing the scores was also compounded with each iteration of updating the scoring system to a newer version. Yet the latest scoring systems still demonstrate unsatisfactory performance, especially when applied in a new setting. One relative advantage of using a data-driven model like CTS-RNN or HCR is that the process for developing such a prediction model can be a \textbf{retrospective} study; it would not require conducting an expensive study with complex selection criteria and regulatory compliance protocols.

For models utilizing structured features, there is an extensive amount of effort required to extract and clean the data. The code for extraction and preprocessing to the re-create the feature set of \citet{Harutyunyan2017-hy} consists of over 2000 lines in multiple files and takes a considerable amount of time. The variables were manually selected, re-sampled to a common period resulting in a loss of temporal information and missing values were imputed which can be problematic. Pragmatically, the HCR approach is much easier to implement because it only requires notes as an input as opposed to the re-sampled collection of physiological variables. Notes are routinely compiled in patient encounters and would require at most a transcriber or an OCR software for digitizing, whereas clinical time series require specialized, expensive and sometimes invasive equipment for acquisition. The development of an HCR approach would not require immense extraction \& processing of multivariate data and has a lower risk of being biased. In the context of actual practical implementations, this presents a trade-off between preprocessing efforts and model complexity \& training time.

Completely ignoring the rich information present in physiological time series would be ignorant though. The HCR has demonstrated that it can be used as a part of a larger multi-modal setup. Any other modalities could be processed by different sequences of layers to produce feature vectors that can be concatenated with the output of the temporal module and hence, create a more informative `patient vector' for classification. For example, convolutional layers could be used to encode medical images into a latent vector. Static data such as demographic data could be encoded by fully-connected layer(s) or just concatenated with the patient vector before classification.

Moreover, the HCR, with other deep learning approaches, is highly capable of being adapted to different situations through {transfer learning}. It can be pre-trained on a public dataset like MIMIC and then fine-tuned on the prospective organization's private data, making extendibility easier. This is useful especially for hospitals with less data: \citet{Desautels2017-wt} tested transfer learning for mortality prediction by training a boosted ensemble of decision trees on the MIMIC-III dataset with different proportions of private data from UCSF (UCSF was the destination where the model was to be applied). It was found that by using MIMIC-III in combination rather than only the private UCSF data, the amount of training data required to surpass 0.80 AUROC decreases from more than 4000 patients to fewer than 220. Moreover, it decreases the clinical data collection time from approximately 6 months to less than 10 days.

One major shortcoming of the HCR architecture is the lack of interpretability. It is difficult to narrow down what part of the notes is affecting the predicted mortality. The black-box nature of the setup doesn't allow for detecting any relationships that might exist between the patient's notes and the predicted mortality. However, \citet{Caicedo-Torres2019-hg} were able to add some interpretability to their convolutional neural network that used clinical time series as an input with DeepLIFT \citep{Shrikumar2017-ss}. Testing interpretability methods on HCR and other notes-based neural networks for mortality prediction seems a very viable direction for future work.

\section{Conclusion}
A study was performed on quantifying the performance of using unstructured free-text notes in the task of mortality prediction with comparison to existing approaches in practice and research. A hierarchical architecture was conceptualized for predicting mortality from patient notes compiled in the initial period of their stay. It utilizes the unstructured format of free text which ensures that all the data available on the patient is used in the prediction rather than limiting to a fixed and pre-determined set of physiological variables. 

It was found that the proposed approach of using notes to predict mortality performed better than severity scores and clinical time series on the same cohort. Augmenting the notes with clinical time series features led to higher average metrics. This shows the value and predictive power of notes for clinical prediction and furthermore demonstrates the need to incorporate more raw unstructured data into existing clinical prediction approaches. 

There are plenty of directions for future research in this approach. The hierarchical model could be improved by using recent developments in text modeling such as implementing the temporal module as another architecture that can keep track of temporality better \textit{e.g.} a transformer \citep{Vaswani2017-ga}. The semantical module could additionally utilize metadata about the notes such as the type of note or author and the explicit time of charting. As mentioned earlier, different methods of interpretability could be explored on the model to have a more nuanced understanding of the relationships between the text and the predicted mortality.

\bibliography{outcomes}

\begin{thebibliography}{48}
\providecommand{\natexlab}[1]{#1}
\providecommand{\url}[1]{\texttt{#1}}
\expandafter\ifx\csname urlstyle\endcsname\relax
  \providecommand{\doi}[1]{doi: #1}\else
  \providecommand{\doi}{doi: \begingroup \urlstyle{rm}\Url}\fi

\bibitem[Aggarwal et~al.(2006)Aggarwal, Sarkar, Gupta, and
  Jindal]{Aggarwal2006-ay}
A.~N. Aggarwal, P.~Sarkar, D.~Gupta, and S.~K. Jindal.
\newblock Performance of standard severity scoring systems for outcome
  prediction in patients admitted to a respiratory intensive care unit in north
  india.
\newblock \emph{Respirology}, 11\penalty0 (2):\penalty0 196--204, Mar. 2006.
\newblock ISSN 1323-7799.
\newblock \doi{10.1111/j.1440-1843.2006.00828.x}.

\bibitem[Bahadori and Lipton(2019)]{Bahadori2019-qx}
M.~T. Bahadori and Z.~C. Lipton.
\newblock {Temporal-Clustering Invariance in Irregular Healthcare Time Series}.
\newblock Apr. 2019.

\bibitem[Beaulieu-Jones(2017)]{Beaulieu2017-zw}
B.~K. Beaulieu-Jones.
\newblock {Machine Learning for Structured Clinical Data}.
\newblock July 2017.

\bibitem[Boag et~al.(2018)Boag, Doss, Naumann, and Szolovits]{Boag2018-mw}
W.~Boag, D.~Doss, T.~Naumann, and P.~Szolovits.
\newblock What's in a note? unpacking predictive value in clinical note
  representations.
\newblock \emph{AMIA Jt Summits Transl Sci Proc}, 2017:\penalty0 26--34, May
  2018.
\newblock ISSN 2153-4063.

\bibitem[Bojanowski et~al.(2017)Bojanowski, Grave, Joulin, and
  Mikolov]{bojanowski2016enriching}
P.~Bojanowski, E.~Grave, A.~Joulin, and T.~Mikolov.
\newblock Enriching word vectors with subword information.
\newblock \emph{Transactions of the Association for Computational Linguistics},
  5:\penalty0 135--146, 2017.

\bibitem[Brown and Crede(1995)]{Brown1995-bn}
M.~C. Brown and W.~B. Crede.
\newblock {Predictive ability of acute physiology and chronic health evaluation
  II scoring applied to human immunodeficiency virus-positive patients}.
\newblock \emph{Critical care medicine}, 23\penalty0 (5):\penalty0 848--853,
  May 1995.
\newblock ISSN 0090-3493.

\bibitem[Caicedo-Torres and Gutierrez(2019)]{Caicedo-Torres2019-hg}
W.~Caicedo-Torres and J.~Gutierrez.
\newblock {ISeeU: Visually interpretable deep learning for mortality prediction
  inside the ICU}.
\newblock Jan. 2019.

\bibitem[Cheng(2017)]{Cheng2017-ma}
J.~Y. Cheng.
\newblock Mortality prediction in status epilepticus with the {APACHE} {II}
  score.
\newblock \emph{Pediatr. Crit. Care Med.}, 18\penalty0 (4):\penalty0 310--317,
  Nov. 2017.
\newblock ISSN 1529-7535, 1751-1437.
\newblock \doi{10.1177/1751143717715967}.

\bibitem[Cho et~al.(2014)Cho, van Merrienboer, Gulcehre, Bahdanau, Bougares,
  Schwenk, and Bengio]{Cho2014-py}
K.~Cho, B.~van Merrienboer, C.~Gulcehre, D.~Bahdanau, F.~Bougares, H.~Schwenk,
  and Y.~Bengio.
\newblock {Learning Phrase Representations using RNN Encoder-Decoder for
  Statistical Machine Translation}.
\newblock In \emph{{Proceedings of the 2014 Conference on Empirical Methods in
  Natural Language Processing (EMNLP)}}, pages 1724--1734, June 2014.

\bibitem[Chollet et~al.(2015)]{chollet2015keras}
F.~Chollet et~al.
\newblock Keras.
\newblock \url{https://keras.io}, 2015.

\bibitem[Desautels et~al.(2017)Desautels, Calvert, Hoffman, Mao, Jay, Fletcher,
  Barton, Chettipally, Kerem, and Das]{Desautels2017-wt}
T.~Desautels, J.~Calvert, J.~Hoffman, Q.~Mao, M.~Jay, G.~Fletcher, C.~Barton,
  U.~Chettipally, Y.~Kerem, and R.~Das.
\newblock {Using Transfer Learning for Improved Mortality Prediction in a
  Data-Scarce Hospital Setting}.
\newblock \emph{Biomedical informatics insights}, 9, June 2017.
\newblock ISSN 1178-2226.
\newblock \doi{10.1177/1178222617712994}.

\bibitem[F{\'e}ry-Lemonnier et~al.(1995)F{\'e}ry-Lemonnier, Landais, Loirat,
  Kleinknecht, and Brivet]{Fery-Lemonnier1995-ut}
E.~F{\'e}ry-Lemonnier, P.~Landais, P.~Loirat, D.~Kleinknecht, and F.~Brivet.
\newblock Evaluation of severity scoring systems in {ICUs---translation},
  conversion and definition ambiguities as a source of inter-observer
  variability in {APACHE II}, {SAPS} and {OSF}.
\newblock \emph{Intensive Care Med.}, 21\penalty0 (4):\penalty0 356--360, Apr.
  1995.
\newblock ISSN 0342-4642, 1432-1238.
\newblock \doi{10.1007/BF01705416}.

\bibitem[Ghassemi et~al.(2014)Ghassemi, Naumann, Doshi-Velez, Brimmer, Joshi,
  Rumshisky, and Szolovits]{Ghassemi2014-qc}
M.~Ghassemi, T.~Naumann, F.~Doshi-Velez, N.~Brimmer, R.~Joshi, A.~Rumshisky,
  and P.~Szolovits.
\newblock Unfolding physiological state: Mortality modelling in intensive care
  units.
\newblock \emph{KDD}, 2014:\penalty0 75--84, Aug. 2014.
\newblock ISSN 2154-817X.
\newblock \doi{10.1145/2623330.2623742}.

\bibitem[Grnarova et~al.(2016)Grnarova, Schmidt, Hyland, and
  Eickhoff]{Grnarova2016-pj}
P.~Grnarova, F.~Schmidt, S.~L. Hyland, and C.~Eickhoff.
\newblock Neural document embeddings for intensive care patient mortality
  prediction.
\newblock Dec. 2016.

\bibitem[Haniffa et~al.(2018)Haniffa, Isaam, De~Silva, Dondorp, and
  De~Keizer]{Haniffa2018-mi}
R.~Haniffa, I.~Isaam, A.~P. De~Silva, A.~M. Dondorp, and N.~F. De~Keizer.
\newblock Performance of critical care prognostic scoring systems in low and
  middle-income countries: a systematic review.
\newblock \emph{Crit. Care}, 22\penalty0 (1):\penalty0 18, Jan. 2018.
\newblock ISSN 0270-7462.
\newblock \doi{10.1186/s13054-017-1930-8}.

\bibitem[Harutyunyan et~al.(2017)Harutyunyan, Khachatrian, Kale, and
  Galstyan]{Harutyunyan2017-hy}
H.~Harutyunyan, H.~Khachatrian, D.~C. Kale, and A.~Galstyan.
\newblock Multitask learning and benchmarking with clinical time series data.
\newblock Mar. 2017.

\bibitem[Johnson(2014)]{johnson2014mortality}
A.~E. Johnson.
\newblock \emph{Mortality prediction and acuity assessment in critical care}.
\newblock PhD thesis, University of Oxford, 2014.

\bibitem[Johnson and Mark(2017)]{johnson2017real}
A.~E. Johnson and R.~G. Mark.
\newblock Real-time mortality prediction in the intensive care unit.
\newblock In \emph{AMIA Annual Symposium Proceedings}, volume 2017, page 994.
  American Medical Informatics Association, 2017.

\bibitem[Johnson et~al.(2013)Johnson, Kramer, and Clifford]{johnson2013new}
A.~E. Johnson, A.~A. Kramer, and G.~D. Clifford.
\newblock A new severity of illness scale using a subset of acute physiology
  and chronic health evaluation data elements shows comparable predictive
  accuracy.
\newblock \emph{Critical care medicine}, 41\penalty0 (7):\penalty0 1711--1718,
  2013.

\bibitem[Johnson et~al.(2016)Johnson, Pollard, Shen, Li-wei, Feng, Ghassemi,
  Moody, Szolovits, Celi, and Mark]{johnson2016mimic}
A.~E. Johnson, T.~J. Pollard, L.~Shen, H.~L. Li-wei, M.~Feng, M.~Ghassemi,
  B.~Moody, P.~Szolovits, L.~A. Celi, and R.~G. Mark.
\newblock Mimic-iii, a freely accessible critical care database.
\newblock \emph{Scientific data}, 3:\penalty0 160035, 2016.

\bibitem[Kingma and Ba(2015)]{Kingma2015-mc}
D.~Kingma and J.~Ba.
\newblock Adam: A method for stochastic optimization.
\newblock In \emph{International Conference on Learning Representations}, 2015.

\bibitem[Knaus et~al.(1991)Knaus, Wagner, Draper, Zimmerman, Bergner, Bastos,
  Sirio, Murphy, Lotring, Damiano, et~al.]{knaus1991apache}
W.~A. Knaus, D.~P. Wagner, E.~A. Draper, J.~E. Zimmerman, M.~Bergner, P.~G.
  Bastos, C.~A. Sirio, D.~J. Murphy, T.~Lotring, A.~Damiano, et~al.
\newblock The apache iii prognostic system: risk prediction of hospital
  mortality for critically iii hospitalized adults.
\newblock \emph{Chest}, 100\penalty0 (6):\penalty0 1619--1636, 1991.

\bibitem[Kocbek et~al.(2017)Kocbek, Fija{\v c}ko, Zorman, Kocbek, and {\v
  S}tiglic]{Kocbek2017-wm}
P.~Kocbek, N.~Fija{\v c}ko, M.~Zorman, S.~Kocbek, and G.~{\v S}tiglic.
\newblock Improving mortality prediction for intensive care unit patients using
  text mining techniques.
\newblock In \emph{Proceedings of {SiKDD} 2017 Conference on Data Mining and
  Data Warehouses}, 2017.

\bibitem[Krishnan and Kamath(2018)]{Krishnan2018-cq}
G.~S. Krishnan and S.~S. Kamath.
\newblock A supervised learning approach for {ICU} mortality prediction based
  on unstructured electrocardiogram text reports.
\newblock In \emph{Natural Language Processing and Information Systems}, pages
  126--134. Springer International Publishing, 2018.
\newblock \doi{10.1007/978-3-319-91947-8\_13}.

\bibitem[Le~Gall et~al.(1993)Le~Gall, Lemeshow, and Saulnier]{le1993new}
J.-R. Le~Gall, S.~Lemeshow, and F.~Saulnier.
\newblock A new simplified acute physiology score (saps ii) based on a
  european/north american multicenter study.
\newblock \emph{Jama}, 270\penalty0 (24):\penalty0 2957--2963, 1993.

\bibitem[Lehman et~al.(2012)Lehman, Saeed, Long, Lee, and Mark]{Lehman2012-jk}
L.-W. Lehman, M.~Saeed, W.~Long, J.~Lee, and R.~Mark.
\newblock Risk stratification of {ICU} patients using topic models inferred
  from unstructured progress notes.
\newblock \emph{AMIA Annu. Symp. Proc.}, 2012:\penalty0 505--511, Nov. 2012.
\newblock ISSN 1942-597X, 1559-4076.

\bibitem[Murdoch and Detsky(2013)]{MurdochBigH80}
T.~B. Murdoch and A.~S. Detsky.
\newblock {The Inevitable Application of Big Data to Health Care}.
\newblock \emph{JAMA}, 309\penalty0 (13):\penalty0 1351--1352, 04 2013.
\newblock ISSN 0098-7484.
\newblock \doi{10.1001/jama.2013.393}.

\bibitem[Pappachan et~al.(1999)Pappachan, Millar, Bennett, and
  Smith]{Pappachan1999-ep}
J.~V. Pappachan, B.~Millar, E.~D. Bennett, and G.~B. Smith.
\newblock Comparison of outcome from intensive care admission after adjustment
  for case mix by the {APACHE} {III} prognostic system.
\newblock \emph{Chest}, 115\penalty0 (3):\penalty0 802--810, Mar. 1999.
\newblock ISSN 0012-3692.

\bibitem[Purushotham et~al.(2018)Purushotham, Meng, Che, and
  Liu]{Purushotham2017-sf}
S.~Purushotham, C.~Meng, Z.~Che, and Y.~Liu.
\newblock {Benchmarking deep learning models on large healthcare datasets}.
\newblock \emph{Journal of biomedical informatics}, 83:\penalty0 112--134, July
  2018.
\newblock ISSN 1532-0464, 1532-0480.
\newblock \doi{10.1016/j.jbi.2018.04.007}.

\bibitem[Reddi et~al.(2018)Reddi, Kale, and Kumar]{amsgrad2018on}
S.~J. Reddi, S.~Kale, and S.~Kumar.
\newblock On the convergence of adam and beyond.
\newblock In \emph{International Conference on Learning Representations}, 2018.

\bibitem[Resnik et~al.(2008)Resnik, Niv, Nossal, Kapit, and
  Toren]{Resnik2008-nt}
P.~Resnik, M.~Niv, M.~Nossal, A.~Kapit, and R.~Toren.
\newblock {Communication of clinically relevant information in electronic
  health records: a comparison between structured data and unrestricted
  physician language}.
\newblock \emph{Perspectives in health information management}, 2008.

\bibitem[Russell(2015)]{Russellbook-pfcc4}
J.~A. Russell.
\newblock Assessment of severity of illness.
\newblock In J.~B. Hall, G.~A. Schmidt, and J.~P. Kress, editors,
  \emph{Principles of Critical Care, 4e}, chapter~13, pages 83--96. McGraw-Hill
  Education, New York, NY, 2015.

\bibitem[Sha and Wang(2017)]{Sha2017-ya}
Y.~Sha and M.~D. Wang.
\newblock Interpretable predictions of clinical outcomes with an
  attention-based recurrent neural network.
\newblock In \emph{Proceedings of the 8th {ACM} International Conference on
  Bioinformatics, Computational Biology,and Health Informatics}, ACM-BCB '17,
  pages 233--240, New York, NY, USA, 2017. ACM.
\newblock ISBN 9781450347228.
\newblock \doi{10.1145/3107411.3107445}.

\bibitem[Shickel et~al.(2018)Shickel, Tighe, Bihorac, and
  Rashidi]{Shickel2018-py}
B.~Shickel, P.~J. Tighe, A.~Bihorac, and P.~Rashidi.
\newblock Deep {EHR}: A survey of recent advances in deep learning techniques
  for electronic health record ({EHR}) analysis.
\newblock \emph{IEEE J Biomed Health Inform}, 22\penalty0 (5):\penalty0
  1589--1604, Sept. 2018.
\newblock ISSN 2168-2208, 2168-2194.
\newblock \doi{10.1109/JBHI.2017.2767063}.

\bibitem[Shrikumar et~al.(2017)Shrikumar, Greenside, and
  Kundaje]{Shrikumar2017-ss}
A.~Shrikumar, P.~Greenside, and A.~Kundaje.
\newblock {Learning Important Features Through Propagating Activation
  Differences}.
\newblock Apr. 2017.

\bibitem[Si and Roberts(2019)]{Si2019-ru}
Y.~Si and K.~Roberts.
\newblock {Deep Patient Representation of Clinical Notes via Multi-Task
  Learning for Mortality Prediction}.
\newblock \emph{AMIA Joint Summits on Translational Science proceedings. AMIA
  Joint Summits on Translational Science}, 2019:\penalty0 779--788, May 2019.
\newblock ISSN 2153-4063.

\bibitem[Song et~al.(2018)Song, Rajan, Thiagarajan, and Spanias]{Song2018-ag}
H.~Song, D.~Rajan, J.~J. Thiagarajan, and A.~Spanias.
\newblock Attend and diagnose: Clinical time series analysis using attention
  models.
\newblock In \emph{Proceedings of the Thirty-Second {AAAI} Conference on
  Artificial Intelligence, New Orleans, Louisiana, USA, February 2-7, 2018},
  Apr. 2018.

\bibitem[Suistomaa et~al.(2000)Suistomaa, Kari, Ruokonen, and
  Takala]{Suistomaa2000-lv}
M.~Suistomaa, A.~Kari, E.~Ruokonen, and J.~Takala.
\newblock {Sampling rate causes bias in APACHE II and SAPS II scores}.
\newblock \emph{Intensive care medicine}, 26\penalty0 (12):\penalty0
  1773--1778, Dec. 2000.
\newblock ISSN 0342-4642.

\bibitem[Suresh et~al.(2018)Suresh, Gong, and Guttag]{Suresh2018-am}
H.~Suresh, J.~J. Gong, and J.~V. Guttag.
\newblock Learning tasks for multitask learning: Heterogenous patient
  populations in the icu.
\newblock In \emph{Proceedings of the 24th ACM SIGKDD International Conference
  on Knowledge Discovery \& Data Mining}, KDD '18, pages 802--810, New York,
  NY, USA, 2018. ACM.
\newblock ISBN 978-1-4503-5552-0.
\newblock \doi{10.1145/3219819.3219930}.

\bibitem[Sushil et~al.(2018)Sushil, {\v S}uster, Luyckx, and
  Daelemans]{Sushil2017-hj}
M.~Sushil, S.~{\v S}uster, K.~Luyckx, and W.~Daelemans.
\newblock {Patient representation learning and interpretable evaluation using
  clinical notes}.
\newblock \emph{Journal of biomedical informatics}, 84:\penalty0 103--113, Aug.
  2018.
\newblock ISSN 1532-0464, 1532-0480.
\newblock \doi{10.1016/j.jbi.2018.06.016}.

\bibitem[Tompson et~al.(2014)Tompson, Goroshin, Jain, LeCun, and
  Bregler]{Tompson2014-lf}
J.~Tompson, R.~Goroshin, A.~Jain, Y.~LeCun, and C.~Bregler.
\newblock {Efficient Object Localization Using Convolutional Networks}.
\newblock Nov. 2014.

\bibitem[Tran and Lee(2018)]{Tran2018-ti}
N.~Tran and J.~Lee.
\newblock Using multiple sentiment dimensions of nursing notes to predict
  mortality in the intensive care unit.
\newblock In \emph{2018 {IEEE} {EMBS} International Conference on Biomedical
  Health Informatics ({BHI})}, pages 283--286, Mar. 2018.
\newblock \doi{10.1109/BHI.2018.8333424}.

\bibitem[Vaswani et~al.(2017)Vaswani, Shazeer, Parmar, Uszkoreit, Jones, Gomez,
  Kaiser, and Polosukhin]{Vaswani2017-ga}
A.~Vaswani, N.~Shazeer, N.~Parmar, J.~Uszkoreit, L.~Jones, A.~N. Gomez,
  {\L}.~U. Kaiser, and I.~Polosukhin.
\newblock {Attention is All you Need}.
\newblock In I.~Guyon, U.~V. Luxburg, S.~Bengio, H.~Wallach, R.~Fergus,
  S.~Vishwanathan, and R.~Garnett, editors, \emph{{Advances in Neural
  Information Processing Systems 30}}, pages 5998--6008. Curran Associates,
  Inc., 2017.

\bibitem[Waudby-Smith et~al.(2018)Waudby-Smith, Tran, Dubin, and
  Lee]{Waudby-Smith2018-cm}
I.~E.~R. Waudby-Smith, N.~Tran, J.~A. Dubin, and J.~Lee.
\newblock Sentiment in nursing notes as an indicator of out-of-hospital
  mortality in intensive care patients.
\newblock \emph{PLoS One}, 13\penalty0 (6):\penalty0 e0198687, June 2018.
\newblock ISSN 1932-6203.
\newblock \doi{10.1371/journal.pone.0198687}.

\bibitem[Weissman et~al.(2018)Weissman, Hubbard, Ungar, and
  {others}]{Weissman2018-vj}
G.~E. Weissman, R.~A. Hubbard, L.~H. Ungar, and {others}.
\newblock Inclusion of unstructured clinical text improves early prediction of
  death or prolonged {ICU} stay.
\newblock \emph{Crit. Care}, 2018.
\newblock ISSN 1364-8535.

\bibitem[Xie et~al.(2017)Xie, Su, Li, Lin, Li, Hu, and Kong]{Xie2017-dm}
J.~Xie, B.~Su, C.~Li, K.~Lin, H.~Li, Y.~Hu, and G.~Kong.
\newblock {A review of modeling methods for predicting in-hospital mortality of
  patients in intensive care unit}.
\newblock \emph{Journal of Emergency and Critical Care Medicine}, 1\penalty0
  (8), Aug. 2017.

\bibitem[Yang et~al.(2016)Yang, Yang, Dyer, He, Smola, and Hovy]{Yang2016-cr}
Z.~Yang, D.~Yang, C.~Dyer, X.~He, A.~Smola, and E.~Hovy.
\newblock {Hierarchical attention networks for document classification}.
\newblock In \emph{{Proceedings of the 2016 conference of the North American
  chapter of the association for computational linguistics: human language
  technologies}}, pages 1480--1489, 2016.

\bibitem[Zalewski et~al.(2017)Zalewski, Long, Johnson, Mark, and
  Lehman]{Zalewski2017-vz}
A.~Zalewski, W.~Long, A.~E.~W. Johnson, R.~G. Mark, and L.-W.~H. Lehman.
\newblock Estimating patient's health state using latent structure inferred
  from clinical time series and text.
\newblock \emph{IEEE EMBS Int Conf Biomed Health Inform}, 2017:\penalty0
  449--452, Feb. 2017.
\newblock \doi{10.1109/BHI.2017.7897302}.

\end{thebibliography}

\end{document}